\documentclass[runningheads]{llncs}
\usepackage{multicol}
\usepackage{multirow}
\usepackage{graphicx}
\usepackage{amsmath,bm,amssymb}
\usepackage{hyperref}
\usepackage{color}
\usepackage{marvosym}
\usepackage[symbol]{footmisc}

\usepackage[firstpage]{draftwatermark} % 只有第一页加水印
\SetWatermarkText{MICCAI 2023}           % 设置水印内容
\SetWatermarkLightness{0.8}             % 设置水印透明度 0-1
\SetWatermarkScale{3}

\begin{document}

\title{A Reliable and Interpretable Framework of Multi-view Learning for Liver Fibrosis Staging}
\titlerunning{A Reliable and Interpretable Framework of Multi-view Learning}
\author{Zheyao Gao \inst{1}$^\dag$
\and Yuanye Liu\inst{1}$^\dag$
\and Fuping Wu\inst{2} 
\and Nannan Shi\inst{3} 
\and Yuxin Shi\inst{3}
\and Xiahai Zhuang\inst{1}$^{\text{\Letter}}$}
\footnotetext[2]{These two authors contribute equally}

\institute{School of Data Science, Fudan University, Shanghai, China \and Nuffield Department of Population Health, University of Oxford, Oxford, UK \and Department of Radiology, Shanghai Public Health Clinical Center, Shanghai, China. \\
\url{www.sdspeople.fudan.edu.cn/zhuangxiahai/}}
% \author{Anonymous}
% \institute{Anonymous Organization \\
% \email{***@***.***}}

\authorrunning{Anonymous}
\maketitle              % typeset the header of the contribution
\begin{abstract}
Staging of liver fibrosis is important in the diagnosis and treatment planning of patients suffering from liver diseases.
Current deep learning-based methods using abdominal magnetic resonance imaging (MRI) usually take a sub-region of the liver as an input, which nevertheless could miss critical information.
To explore richer representations, we formulate this task as a multi-view learning problem and employ multiple sub-regions of the liver.
Previously, features or predictions are usually combined in an implicit manner, 
and uncertainty-aware methods have been proposed. 
However, these methods could be challenged to capture cross-view representations, 
which can be important in the accurate prediction of staging.
Therefore, we propose a reliable multi-view learning method with interpretable combination rules, which can model global representations to improve the accuracy of predictions.
Specifically, the proposed method estimates uncertainties based on subjective logic to improve reliability, and an explicit combination rule is applied based on Dempster-Shafer's evidence theory with good power of interpretability. Moreover, a data-efficient transformer is introduced to capture representations in the global view.
Results evaluated on enhanced MRI data show that our method delivers superior performance over existing multi-view learning methods.

\end{abstract}
\section{Introduction}
\begin{figure}[t]
    \label{fig_intro}
    \centering 
    \begin{tabular}{ccccc}
        \includegraphics[height=0.25\textwidth]{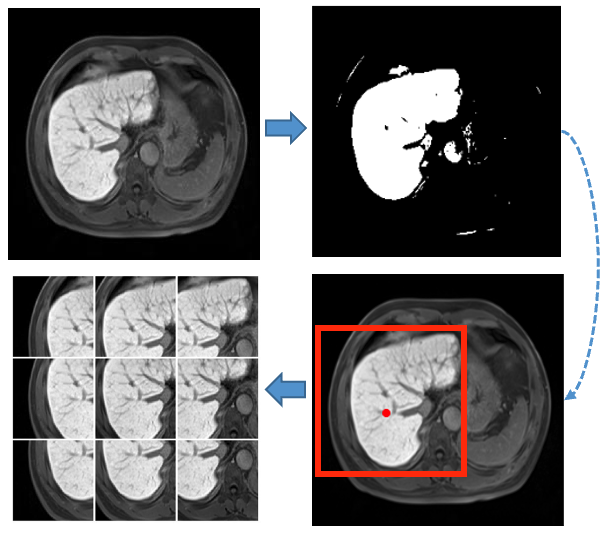} & &\quad &\quad &
        \includegraphics[height=0.25\textwidth]{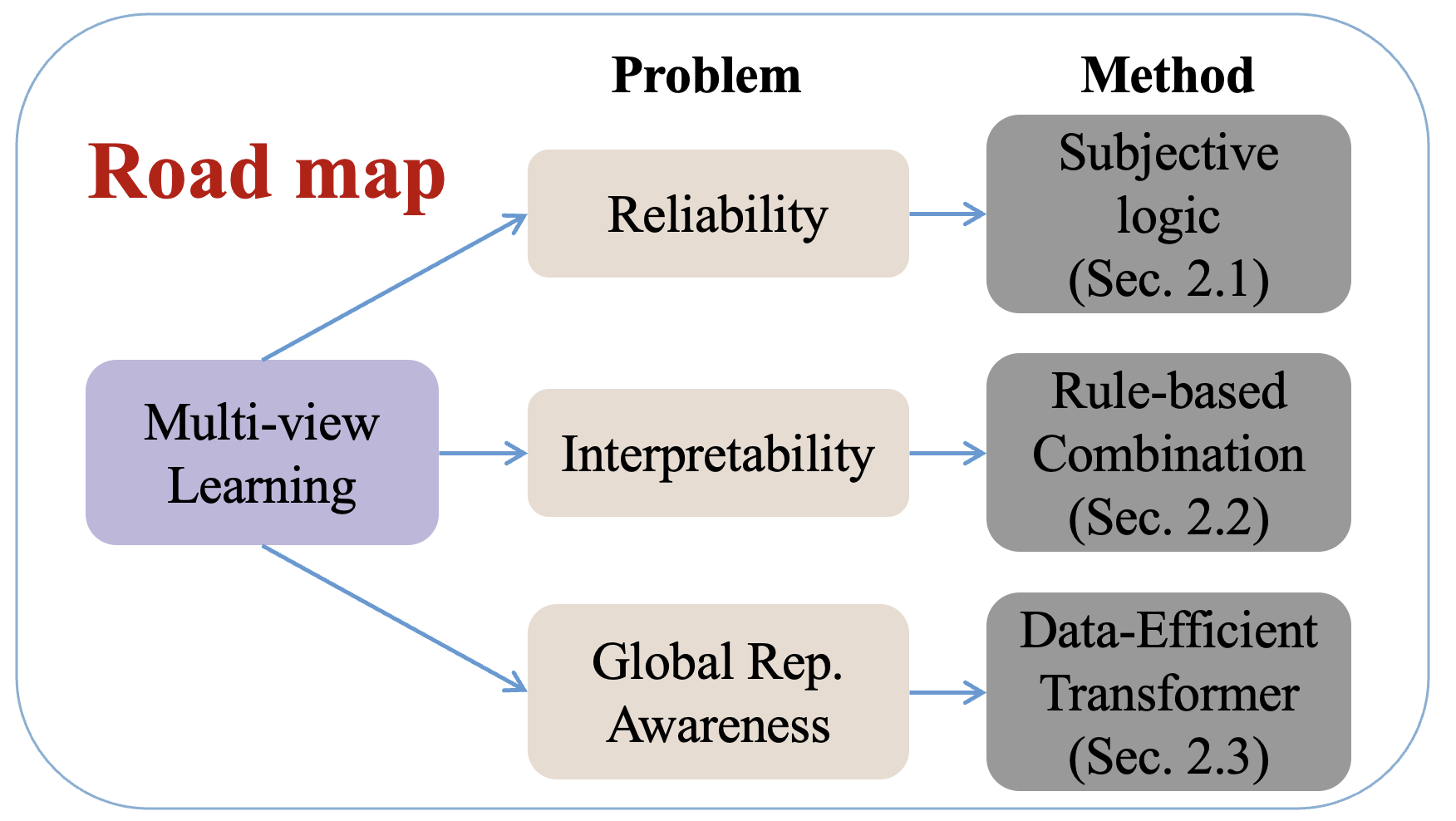}   \\
         (a) & &\quad &\quad\ & (b)
    \end{tabular}

    \caption{(a) The pipeline to extract sub-views of the liver. First, the foreground is extracted using intensity-based segmentation. Based on the segmentation, a square region of interest (ROI) centered at the centroid of the liver is cropped. Then overlapped sliding windows are used in the ROI to obtain nine sub-views of the liver. (b) The road map of this work.}
\vspace{-0.8em}
\end{figure}

Viral or metabolic chronic liver diseases that cause liver fibrosis impose great challenges on global health. Accurate staging for the severity of liver fibrosis is essential in the diagnosis of various liver diseases. 
Current deep learning-based methods \cite{liverCT,liverR} mainly use abdominal MRI and computed tomography (CT) data for liver fibrosis staging. 
Usually, a square sub-region of the liver instead of the whole image is cropped as input features, since the shape of the liver is irregular and unrelated anatomies in the abdominal image could disturb the training of deep learning models. 
To automatically extract the region of interest (ROI), a recent work \cite{liverER} proposes to use slide windows to crop multiple image patches around the centroid of the liver for data augmentation. However, it only uses one patch as input at each time, which only captures a sub-view of the liver. To exploit informative features across the whole liver, we formulate this task as a multi-view learning problem and consider each patch as a view. The pipeline for view extraction is shown in Fig. \ref{fig_intro}(a). A square region of interest (ROI) is cropped based on the segmentation of the foreground. Then nine sub-views of the liver are extracted in the ROI through overlapped sliding windows.

The aim of multi-view learning is to exploit complementary information from multiple features \cite{multiview}.
The central problem is how to integrate features from multiple views properly.
In addition to the naive method that concatenates features at the input level \cite{input}, feature-level fusion strategies seek a common representation between different views through canonical correlation analysis \cite{CCA2,CCA1} or maximizing the mutual information between different views using contrastive learning \cite{contrast2,contrast1}.
In terms of decision-level fusion, the widely used methods are decision averaging \cite{decision1}, decision voting \cite{decision2}, and attention-based decision fusion \cite{decision3}.
However, in the methods above, the weighting of multi-view features is either equal or learned implicitly through model training, which undermines the interpretability of the decision-making process. 
Besides, they are not capable of quantifying uncertainties, which could be non-trustworthy in healthcare applications.

To enhance the interpretability and reliability of multi-view learning methods, recent works have proposed uncertainty-aware decision-level fusion strategies.
Typically, they first estimate uncertainties through Bayesian methods such as Monte-Carlo dropout \cite{Uncertainty2}, variational inference \cite{uncertainty1}, ensemble methods \cite{masksembles}, and evidential learning \cite{evidential}.
Then, the predictions from each view are aggregated through explicit uncertainty-aware combination rules \cite{rule2,rule1}, as logic rules are commonly acknowledged to be interpretable in a complex model \cite{interpretable}.
However, the predictions before the combination are made based on each independent view. Cross-view features are not captured to support the final prediction.
In our task, global features could also be informative in the staging of liver fibrosis.

In this work, we propose an uncertainty-aware multi-view learning method with an interpretable fusion strategy of liver fibrosis staging, which captures both global features across views and local features in each independent view. The road map for this work is shown in Fig. \ref{fig_intro}(b).
The uncertainty of each view is estimated through the evidential network and subjective logic to improve reliability. 
Based on the uncertainties, we apply an explicit combination rule according to Dempster-Shafer's evidence theory to obtain the final prediction, which improves explainability.
Moreover, we incorporate an additional global view to model the cross-view representation through the data-efficient transformer.

Our contribution has three folds. First, we are the first to formulate liver fibrosis staging as a multi-view learning problem and propose an uncertainty-aware framework with an interpretable fusion strategy based on Dempster-Shafer Evidence Theory. 
Second, we propose to incorporate global representation in the multi-view learning framework through the data-efficient transformer network.
Third, we evaluate the proposed framework on enhanced liver MRI data. The results show that our method outperforms existing multi-view learning methods and yields lower calibration errors than other uncertainty estimation methods.

\begin{figure}[t]
    \label{pipeline}
    \centering 
    \includegraphics[width=\textwidth]{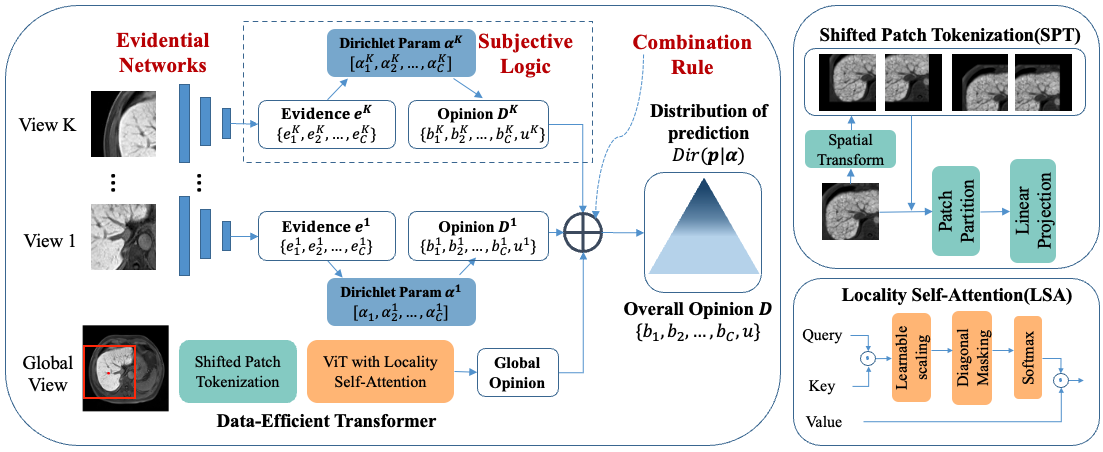} 
    \caption{The left side shows the main framework. Multi-view images are first encoded as evidence vectors by evidential networks. For each view, an opinion with uncertainty $u$ is derived from evidence, under the guidance of subjective logic. Finally, the opinions are combined based on an explicit rule to derive the overall opinion, which can be converted to the distribution of classification probabilities. The right side illustrates the SPT and LSA modules in the data-efficient transformer that serves as the evidential network for the global view.}
    \vspace{-1.0em}
\end{figure}

\section{Methods}
The aim of our method is to derive a distribution of class probabilities with uncertainty based on multiple views of a liver image.
As shown in Fig. \hyperref[pipeline]{2}, our framework mainly consists of three parts, \textit{i.e.}, evidential network, subjective logic, and combination rule. 
The evidential networks encode local views and the whole ROI as global view to evidence vectors $\bm{e}$. For local views, the networks are implemented with the convolutional structure. While for the global view, a data-efficient vision transformer with shifted patch tokenization (SPT) and locality self-attention (LSA) strategy is applied. 
Subjective logic serves as a principle that transforms the vector $\bm{e}$ into the parameter $\bm{\alpha}$ of the Dirichlet distribution of classification predictions, and the opinion $\bm{D}$ with uncertainty $u$.
Then, Dempster’s combination rule is applied to form the final opinion with overall uncertainty, which can be transformed into the final prediction.
The details of subjective logic, Dempster's combination rule, the data-efficient transformer, and the training paradigm are discussed in the following sections.

\subsection{Subjective Logic for Uncertainty Estimation}

Subjective logic, as a generalization of the Bayesian theory, is a principled method of probabilistic reasoning under uncertainty \cite{subjective}. It serves as the guideline of the estimation of both uncertainty and distribution of predicted probabilities in our framework.
Given an image $x_k$ from view $k$, $k\in\{1,2,\cdots,K\}$, the evidence vector $\bm{e}^k=[e_1^k,e_2^k,...,e_C^k]$ with non-negative elements for $C$ classes is estimated through the evidential network, which is implemented using a classification network with softplus activation for the output.

According to subjective logic, the Dirichlet distribution of class probabilities $Dir(\bm{p}^k|{\bm{\alpha}^k)}$ is determined by the evidence. For simplicity, we follow \cite{evidential} and derive the parameter of the distribution by $\bm{\alpha}^k=\bm{e}^k+1$. Then the Dirichlet distribution is mapped to an opinion $\bm{D}^k=\{\{b_c^k\}_{c=1}^C,u^k\}$, subject to
\vspace{-0.5em}
% Based on the evidence, an opinion $\bm{D}^k=\{\{b_c^k\}_{c=1}^C,u^k\}$ with uncertainty could be derived, subject to,
\begin{equation}
    u^k+\sum_{c=1}^C{b_c^k}=1,
\end{equation}\\[-0.4cm]
where $b_c^k=\frac{\alpha_c^k-1}{S^k}$ is the belief mass for class $c$, $S^k=\sum_{c=1}^C{\alpha_c^k}$ is the Dirichlet strength, and $u^k=\frac{C}{S^k}$ indicates the uncertainty.

The predicted probabilities $\tilde{\bm{p}}^k\in \mathbb{R}^C$ of all classes are the expectation of Dirichlet distribution, \textit{i.e.}, $\tilde{\bm{p}}^k=\mathbb{E}_{Dir(\bm{p}^k|\bm{\alpha}^k)}[\bm{p}^k]$. Therefore, the uncertainty $u^k$ and predicted probabilities $\tilde{\bm{p}}^k$ can be derived in an end-to-end manner.

\subsection{Combination Rule}
Based on opinions derived from each view, Dempster's combination rule \cite{DST} is applied to obtain the overall opinion with uncertainty, which could be converted to the distribution of the final prediction. Specifically, given opinions $\bm{D}^1=\{\{b_c^1\}_{c=1}^C,u^1\}$ and $\bm{D}^2=\{\{b_c^2\}_{c=1}^C,u^2\}$, the combined opinion $\bm{D}=\{\{b_c\}_{c=1}^C,u\}=\bm{D}^1 \oplus \bm{D}^2$ is derived by the following rule,
\begin{equation}\label{rule}
    b_c= \frac{1}{N}(b^1_c b^2_c + b_c^1 u^2 + b_c^2 u^1), u=\frac{1}{N}u^1 u^2,
\end{equation}
where $N=1-\sum_{i\neq j}{b_i^1 b_j^2}$ is the normalization factor. According to Eq. (\ref{rule}), the combination rule indicates that the combined belief $b_c$ depends more on the opinion which is confident (with small $u$). In terms of uncertainty, the combined $u$ is small when at least one opinion is confident.

For opinions from $K$ local views and one global view, the combined opinion could be derived by applying the above rule for $K$ times, \textit{i.e.}, $\bm{D}=\bm{D}^1\oplus\cdots\oplus \bm{D}^K\oplus{\bm{D}^{Global}}$.

\subsection{Global representation modeling}
To capture the global representation, we apply a data-efficient transformer as the evidential network for the global view.
We follow \cite{trans} and improve the performance of the transformer on small datasets by increasing locality inductive bias, \textit{i.e.}, the assumption about relations between adjacent pixels. The standard vision transformer (ViT) \cite{vit} without such assumptions typically require more training data than convolutional networks \cite{bias}. Therefore, we adopt the SPT and LSA strategy to improve the locality inductive bias.

As shown in Fig. \hyperref[pipeline]{2}, SPT is different from the standard tokenization in that the input image is shifted in four diagonal directions by half the patch size, and the shifted images are concatenated with the original images in the channel dimension to further utilize spatial relations between neighboring pixels. Then, the concatenated images are partitioned into patches and linearly projected as visual tokens in the same way as ViT.

LSA modifies self-attention in ViT by sharpening the distribution of the attention map to pay more attention to important visual tokens. As shown in Fig. \hyperref[pipeline]{2}, diagonal masking and temperature scaling are performed before applying softmax to the attention map. Given the input feature $\bm{X}$, The LSA module is formalized as,
\begin{equation}
    L(\bm{X})= \text{softmax}(\mathcal{M}(\bm{qk}^T)/\tau)\bm{v},
\end{equation}
where $\bm{q,k,v}$ are the query, key, and value vectors obtained by linear projections of $\bm{X}$. $\mathcal{M}$ is the diagonal masking operator that sets the diagonal elements of $\bm{qk}^T$ to a small number (\textit{e.g.},$-\infty$). $\tau\in \mathbb{R}$ is the learnable scaling factor.

\subsection{Training Paradigm}
Theoretically, the proposed framework could be trained in an end-to-end manner. For each view $k$, we use the integrated cross-entropy loss as in \cite{evidential},\vspace{-0.5em}
\begin{equation}
    \mathcal{L}_{ice}^k=\mathbb{E}_{\bm{p}^k\sim Dir(\bm{p}^k|\bm{\alpha}^k)}[\mathcal{L}_{CE}(\bm{p}^k,\bm{y}^k)]=\sum_{c=1}^Cy_c^k(\psi(S^k)-\psi(\alpha_c^k)),
\end{equation}\\[-0.4cm]
where $\psi$ is the digamma function and $\bm{y}^k$ is the one-hot label. We also apply a regularization term to increase the uncertainty of misclassified samples,
\begin{equation}
    \mathcal{L}^k=\mathcal{L}_{ice}^k+\lambda KL[Dir(\bm{p}^k|\tilde{\bm{\alpha}}^k)||Dir(\bm{p}^k|\bm{1})],
\end{equation}
where $\lambda$ is the balance factor which gradually increases during training and $\tilde{\bm{\alpha}}^k=\bm{y}^k+(1-\bm{y}^k)\odot \bm{\alpha}^k$. The overall loss is the summation of losses from all views and the loss for the combined opinion,\vspace{-0.5em}
\begin{equation}\label{loss}
     \mathcal{L}_{Overall} = \mathcal{L}_{Combined} + \mathcal{L}_{Global} + \sum_{k=1}^K\mathcal{L}^k,
\end{equation}\\[-0.3cm]
where $\mathcal{L}_{Combined}$ and $\mathcal{L}_{Global}$ are losses of the combined and global opinions, implemented in the same way as $\mathcal{L}^k$.
In practice, we pre-train the evidential networks before training with Eq. (\ref{loss}). For local views, we use the model weights pre-trained on ImageNet, and the transformer is pre-trained on the global view images.

\section{Experiments}

\subsection{Dataset}
The proposed method was evaluated on Gd‐EOB‐DTPA-enhanced \cite{liverR} hepatobiliary phase MRI data, including 342 patients acquired from two scanners, \textit{i.e.}, Siemens 1.5T and Siemens 3.0T.
The gold standard was obtained through the pathological analysis of the liver biopsy or liver resection within 3 months before and after MRI scans. Among all patients, 88 individuals were identified with fibrosis stage S1, 41 with S2, 40 with S3, and 174 with the most advanced stage S4. 
Following \cite{liverR}, the slices with the largest liver area in images were selected.
The data were then preprocessed with z-score normalization, resampled to a resolution of $1.5\times 1.5mm^2$, and cropped to $256\times 256$ pixel. For multi-view extraction, the size of the ROI, window, and stride were 160, 96, 32, respectively.

For all experiments, a four-fold cross-validation strategy was employed, and results of two tasks with clinical significance \cite{liverR} were evaluated, \textit{i.e.}, staging cirrhosis (S4 vs S1-3) and identifying substantial fibrosis (S1 vs S2-4). 
To keep a balanced number of samples for each class, we over-sampled the S1 data and under-sampled S4 data in the experiments of staging substantial fibrosis.

\subsection{Implementation details}
Augmentations such as random rescale, flip, and cutout \cite{cutout} were applied during training. We chose ResNet34 as the evidential network for local views. For configurations of the transformer, please refer to supplementary materials. 
The framework was trained using Adam optimizer with an initial learning rate of $1e-4$ for $500$ epochs, which was decreased by using the polynomial scheduler. The balance factor $\lambda$ was set to increase linearly from 0 to 1 during training. The transformer network was pre-trained for 200 epochs using the same setting. The framework was implemented using Pytorch and was run on one Nvidia RTX 3090 GPU.

\subsection{Results}
\noindent \textbf{Comparison with multi-view learning methods}
To assess the effectiveness of the proposed multi-view learning framework for liver fibrosis staging, we compared it with five multi-view learning methods, including Concat \cite{input}, DCCAE \cite{DCCAE}, CMC \cite{contrast1}, PredSum \cite{decision1}, and Attention \cite{decision3}. 
Concat is a commonly used method that concatenates multi-view images at the input level. DCCAE and CMC are feature-level strategies. PredSum and Attention are based on decision-level fusion. Additionally, SingleView \cite{liverER} was adopted as the baseline method for liver fibrosis staging, which uses a single patch as input.
% To assess the effectiveness of the proposed multi-view learning framework for liver fibrosis staging, we compare it with existing multi-view learning methods and the liver fibrosis staging method \cite{liverER} (denoted as SingleView). The naive input-level fusion method (Concat), two feature-level fusion methods (DCCAE, CMC), and two decision-level fusion methods (PredSum, Attention) are compared. 

As shown in Table \hyperref[comparison1]{1}, our method outperformed the SingleView method by 10.3\% and 12\% in AUC on the two tasks, respectively, indicating that the proposed method could exploit more informative features than the method using single view. 
Our method also set the new state of the art, when compared with other multi-view learning methods. 
This could be due to the fact that our method was able to capture both the global and local features, and the uncertainty-aware fusion strategy could be more robust than the methods with implicit fusion strategies.

\begin{table}[t]
\label{comparison1}
    \centering
    \caption{Comparison with multi-view learning methods. Results are evaluated in accuracy (ACC) and area under the receiver operating characteristic curve (AUC) for both tasks.}
    
    % \resizebox{1\textwidth}{!}{
    \setlength{\tabcolsep}{0.02\linewidth}{
    \begin{tabular}{|c|c|c|c|c|}
        \hline
  
      \multirow{2}{*}{Method} &  \multicolumn{2}{c|}{Cirrhosis(S4 vs S1-3)} & \multicolumn{2}{c|}{Substantial Fibrosis(S1 vs S2-4)}\\
      \cline{2-5}
      \multirow{2}{*}{} &ACC&AUC&ACC&AUC\\
      \hline
      SingleView \cite{liverER}  & $77.1\pm3.17$ & $78.7\pm4.17$ & $78.2\pm7.18$ & $75.0\pm11.5$ \\   
      \hline
      Concat \cite{input}        & $80.0\pm2.49$ & $81.8\pm3.17$ & $80.5\pm2.52$ & $83.3\pm3.65$ \\
       \hline
      DCCAE \cite{DCCAE}         & $80.6\pm3.17$ & $82.7\pm4.03$ & $83.1\pm5.30$ & $84.5\pm4.77$ \\
       \hline
      CMC \cite{contrast1}       & $80.6\pm1.95$ & $83.5\pm3.67$ & $83.4\pm3.22$ & $85.3\pm4.06$ \\   
      \hline
      PredSum \cite{decision1}   & $78.8\pm4.16$ & $78.2\pm4.94$ & $81.1\pm2.65$ & $84.9\pm3.21$ \\
      \hline
      Attention \cite{decision3} & $76.2\pm0.98$ & $78.9\pm3.72$ & $81.4\pm4.27$ & $84.4\pm5.34$ \\   
      \hline
      Ours                       & \bm{$84.4\pm1.74$} & \bm{$89.0\pm0.03$} & $\bm{85.5\pm1.91}$ & $\bm{88.4\pm1.84}$  \\
      \hline
    \end{tabular}
    }
    \vspace{-1.0em}
\end{table}

\begin{table}[t] 
    \centering
    \caption{Comparison with uncertainty-aware methods. The expected calibration error (ECE) is evaluated in addition to ACC and AUC. Methods with lower ECE are more reliable.}
    \resizebox{1\textwidth}{!}{
        \begin{tabular}{|c|c|c|c|c|c|c|}
        \hline
        \multirow{2}{*}{Method} &\multicolumn{3}{c|}{Cirrhosis(S4 vs S1-3)} & \multicolumn{3}{c|}{Substantial Fibrosis(S1 vs S2-4)}\\
        \cline{2-7}
        \multirow{2}{*}{} &ACC     & AUC    &ECE&ACC&AUC&ECE\\
        \hline
        Softmax                     & $77.1\pm3.17$ & $78.7\pm4.17$ & $0.256\pm0.040$ & $78.2\pm7.18$ & $83.3\pm3.65$ & $0.237\pm0.065$  \\
        \hline
        % Temp-scaling               & $77.1\pm3.17$ & $78.7\pm4.17$ & $0.218\pm0.042$ & $78.2\pm7.18$ & $75.0\pm11.5$ & $0.215\pm0.072$ \\
        % \hline
        Dropout \cite{Uncertainty2} & $77.1\pm4.89$ & $79.8\pm4.50$ & $0.183\pm0.063$ & $80.2\pm5.00$ & $83.8\pm6.12$ & $0.171\pm0.067$  \\
        \hline
        VI \cite{uncertainty1}      & $77.6\pm2.20$ & $79.5\pm4.50$ & $0.229\pm0.020$ & $81.1\pm2.08$ & $82.2\pm6.12$ & $0.191\pm0.023$  \\
        \hline
        Ensemble \cite{masksembles} & $78.1\pm1.91$ & $80.8\pm3.13$ & $0.181\pm0.040$ & $79.3\pm5.11$ & $80.4\pm3.90$ & $0.193\pm0.031$  \\
        \hline
        Ours                        & $\bm{84.4\pm1.74}$ & $\bm{89.0\pm0.03}$ & $\bm{0.154\pm0.028}$ & $\bm{85.5\pm1.91}$ & $\bm{88.4\pm1.84}$ & $\bm{0.156\pm0.019}$  \\
        \hline
        \end{tabular}
        }
        \label{comparison2}
        \vspace{-0.7em}
\end{table}

\noindent \textbf{Comparison with uncertainty-aware methods.}
To demonstrate reliability, we compared the proposed method with other methods. Specifically, these methods estimate uncertainty using Monte-Carlo dropout (Dropout) \cite{Uncertainty2}, variational inference (VI) \cite{uncertainty1}, ensemble \cite{masksembles}, and softmax entropy \cite{softmax}, respectively.
Following \cite{ECE}, we evaluated the expected calibration error (ECE), which measures the gap between model confidence and expected accuracy. 
% To demonstrate the reliability of the proposed framework, we compared it with other uncertainty-aware
% methods. Following [5], we evaluated the expected calibration error (ECE), which measures the gap between model confidence and expected accuracy. Two uncertainty-aware decision-level fusion methods (Dropout, VI) and the mask ensemble method (Ensemble) based on input-fusion were compared. The results of
% the baseline method that uses softmax prediction as confidence were also shown in Table 2.

% Compared with the uncertainty-ware multi-view learning method that estimates uncertainties via MC dropout and variational inference (VI)
Table \hyperref[comparison2]{2} shows that our method achieved better results in ACC and AUC for both tasks than the other uncertainty-ware multi-view learning methods. 
It indicates that the uncertainty in our framework could paint a clearer picture of the reliability of each view, and thus the final prediction was more accurate based on the proposed scheme of rule-based combination.
Our method also achieved the lowest ECE, indicating that the correspondence between the model confidence and overall results was more accurate.

% \begin{figure}[t]
%     \centering
%     \begin{tabular}{ccccc}
%         \fbox{\includegraphics[height=0.22\textwidth]{figs/uncertainty_illustrate_3.png}} & &\qquad & &
%         \fbox{\includegraphics[height=0.22\textwidth]{figs/uncertainty_illustrate_4.png}}   \\
%         (a) & && & (b)
%     \end{tabular}
%     \caption{Typical samples of stage 4 (a) and stage 1 (b). Visible signs of liver fibrosis are highlighted by circles. Red circles indicate liver fibrosis due to tuberculosis and yellow circles denote liver nodules with high signal intensity. Uncertainties (U) of local and global views estimated by our model were demonstrated. Notably, local views of lower uncertainty contain more signs of fibrosis. Please refer to supplementary materials for more high-resolute images}
%     \label{sample}
% \end{figure}

\begin{figure}[t]
    \centering
        \begin{tabular}{ccccc}
        \fbox{\includegraphics[width=0.45\textwidth]{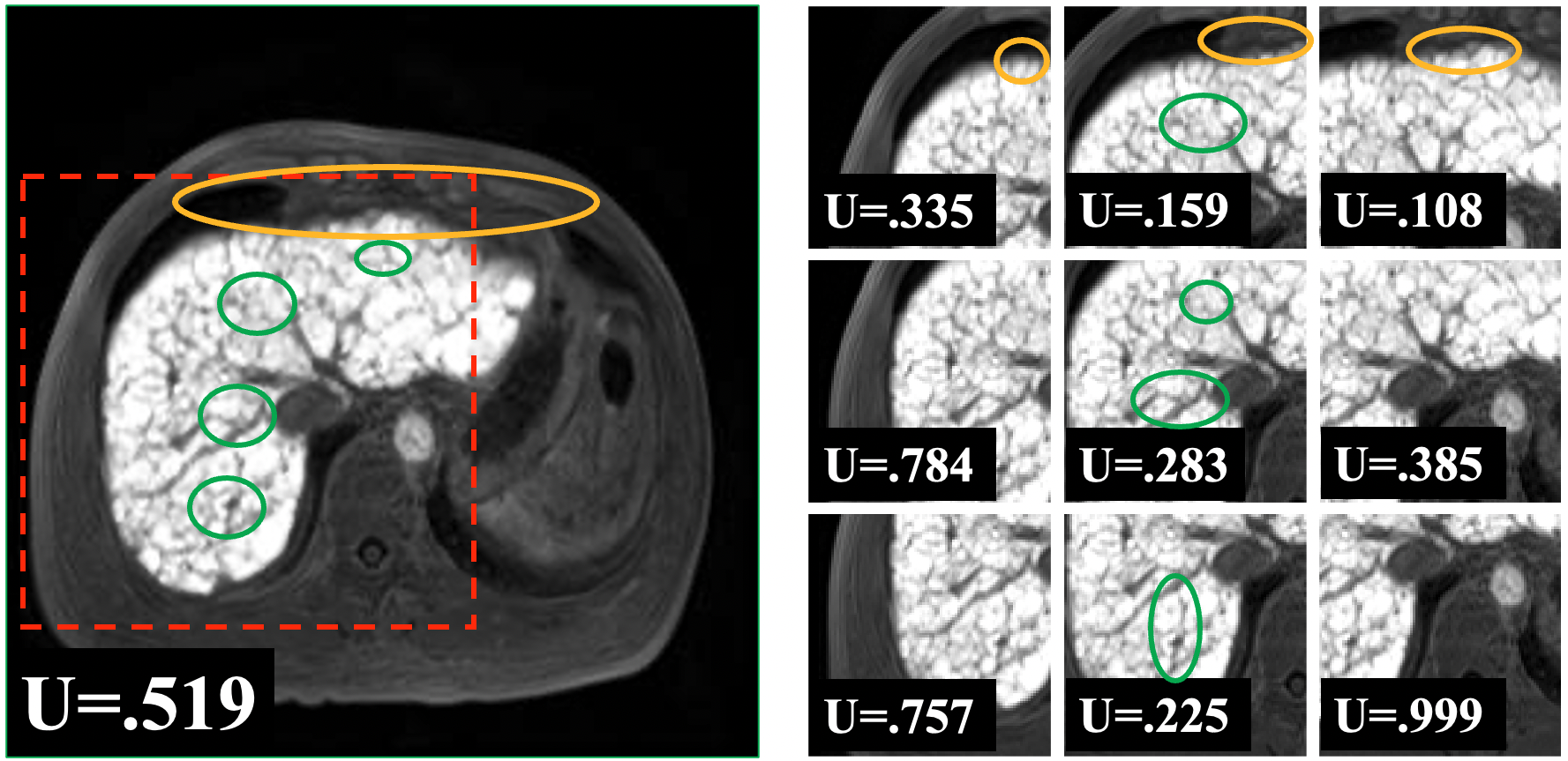}} & &\quad & &
        % & &\qquad & &
        \fbox{\includegraphics[width=0.45\textwidth]{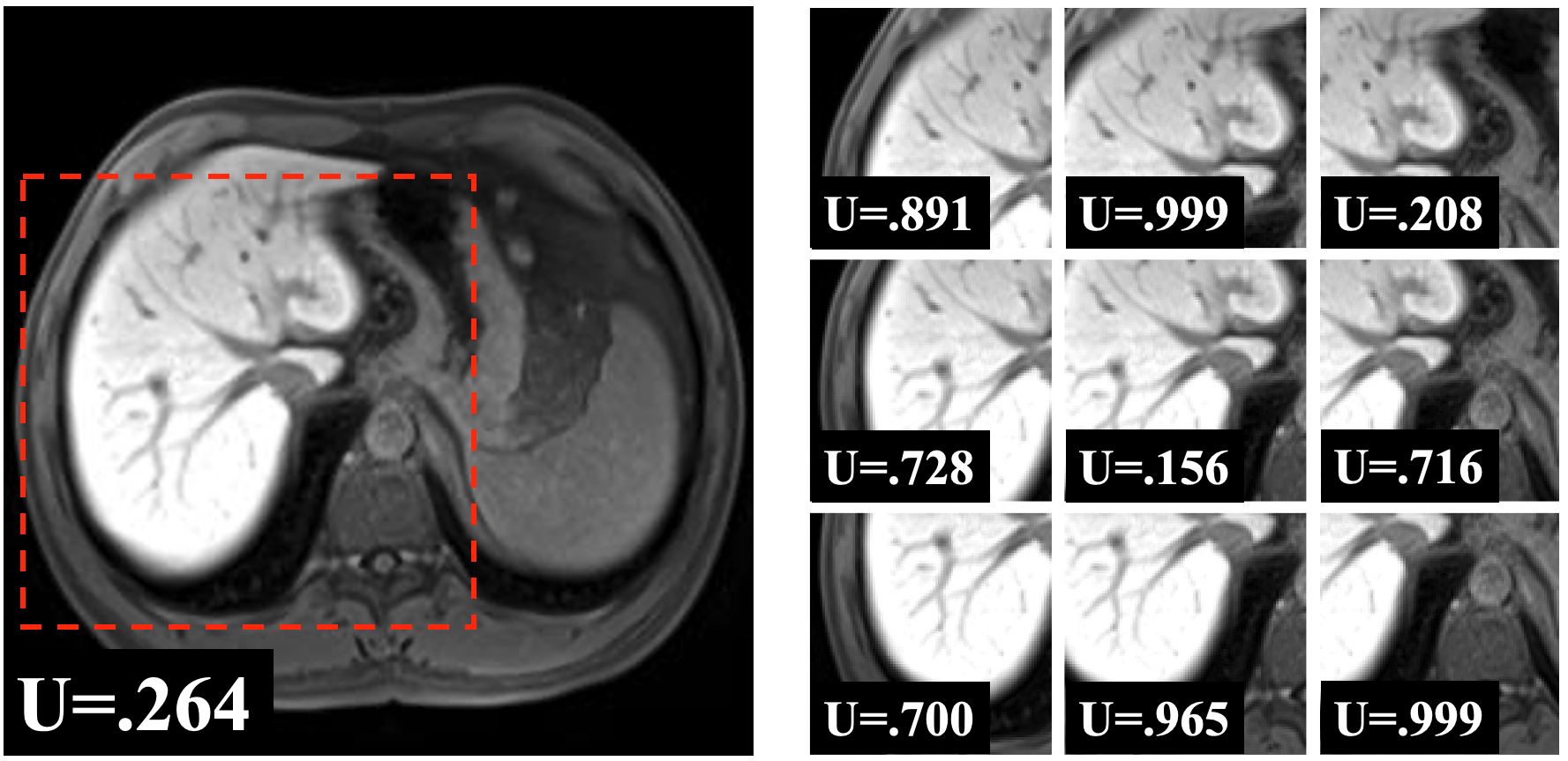}}   \\
         (a) & &\quad & & (b)
    \end{tabular}
    \caption{Typical samples of stage 4 (a) and stage 1 (b). Visible signs of liver fibrosis are highlighted by circles. Yellow circles indicate the nodular surface contour and green circles denote numerous regenerative nodules. Uncertainties (U) of local and global views estimated by our model were demonstrated. Notably, local views of lower uncertainty contain more signs of fibrosis. Please refer to supplementary materials for more high-resolute images}
    \label{sample}
\end{figure}

\begin{table} [t]
    \centering
    \caption{Ablation study for the roles of local and global views, and effectiveness of the data-efficient transformer.}
    \resizebox{1\textwidth}{!}{
    \begin{tabular}{|c|c|c|c|c|c|c|}
    \hline
    \multirow{2}{*}{Method} &  \multicolumn{3}{c|}{Cirrhosis(S4 vs S1-3)} & \multicolumn{3}{c|}{Substantial Fibrosis(S1 vs S2-4)}\\
    \cline{2-7}
    \multirow{2}{*}{} &ACC&AUC&ECE&ACC&AUC&ECE\\
    \hline
    Global View solely   & $76.8\pm2.81$ & $79.4\pm4.76$ & $0.192\pm0.071$ & $82.4\pm3.45$ & $84.9\pm5.42$ & $0.192\pm0.071$                \\
    \hline
    Local Views  solely  & $84.1\pm6.47$ & $88.0\pm8.39$ & $\bm{0.148\pm0.086}$ & $82.0\pm6.07$ & $86.9\pm6.68$ & $0.180\pm0.060$  \\
    \hline
    Both views by CNN     & $82.9\pm3.17$ & $87.8\pm3.09$ & $0.171\pm0.029$ & $82.0\pm3.54$ & $87.1\pm3.47$ & $0.174\pm0.039$  \\
    \hline
    Ours           & $\bm{84.4\pm1.74}$ & $\bm{89.0\pm0.03}$ & $0.154\pm0.028$ & $\bm{85.5\pm1.91}$ & $\bm{88.4\pm1.84}$ & $\bm{0.156\pm0.019}$  \\
    \hline
    \end{tabular}
    }
    \label{ablation}
    \vspace{-1.2em}
\end{table}

\noindent \textbf{Ablation study.} 
We performed this ablation study to investigate the roles of local views and global view, as well as to validate the effectiveness of the data-efficient transformer.

Table \hyperref[ablation]{3} shows that using the global view solely achieved the worst performance in the staging of cirrhosis. This means that it could be difficult to extract useful features without complementary information from local views. 
This is consistent with Fig. \ref{sample}(a), where the uncertainty derived from the global view is high, even if there are many signs of fibrosis. While in Fig. \ref{sample}(b), the uncertainty of the global view is low, which indicates that it is easier to make decisions from the global view when there is no visible sign of fibrosis. 
Therefore, we concluded that the global view was more valuable in identifying substantial fibrosis.
Compared with the method that only used local views, our method gained more improvement in the substantial fibrosis identification task, which further confirms the aforementioned conclusion.
Our method also performed better than the method that applied a convolution neural network (CNN) for the global view. This demonstrates that the proposed data-efficient transformer was more suitable for the modeling of global representation than CNN.

\section{Conclusion}
In this work, we have proposed a reliable and interpretable multi-view learning framework for liver fibrosis staging. Specifically, uncertainty is estimated through subjective logic to improve reliability, and an explicit fusion strategy is applied which promotes interpretability. Furthermore, we use a data-efficient transformer to model the global representation, which improves the performance.

%
% the environments 'definition', 'lemma', 'proposition', 'corollary',
% 'remark', and 'example' are defined in the LLNCS documentclass as well.
%


% \end{thebibliography}
\bibliographystyle{splncs04}
\bibliography{refs}

\end{document}